\documentclass[letterpaper, 10 pt, conference]{ieeeconf}
\IEEEoverridecommandlockouts
\usepackage[space, compress, sort]{cite}
\usepackage{amsmath}
\usepackage{amsfonts}
\usepackage{diagbox}

\usepackage{amsthm}
\usepackage[ruled,vlined]{algorithm2e}
\usepackage{mathtools}
\usepackage{tabularx}
\usepackage{graphicx}
\usepackage{enumerate}
\usepackage{float}
\usepackage{url}
\newcommand{\tabincell}[2]{\begin{tabular}{@{}#1@{}}#2\end{tabular}}

\usepackage{tabularx}
\usepackage{multirow, multicol}
\usepackage{array}
\newcolumntype{P}[1]{>{\centering\arraybackslash}p{#1}}
\newcolumntype{M}[1]{>{\centering\arraybackslash}m{#1}}
\usepackage{booktabs}
\newcolumntype{N}{>{\centering\arraybackslash}m{.5in}}
\newcolumntype{G}{>{\centering\arraybackslash}m{2in}}
\usepackage{verbatim}
\usepackage{booktabs} 
\usepackage{multirow}
\usepackage[dvipsnames]{xcolor}
\usepackage[linkcolor=black,citecolor=black,urlcolor=black,colorlinks=true]{hyperref}
\usepackage{bm}
\usepackage{algpseudocode}

\newcommand{\add}[1]{{{{\color{Black}#1}}}}
\newcommand{\revise}[1]{{{{\color{Black}#1}}}}

\title{\LARGE \bf
Design and Evaluation of Motion Planners for Quadrotors in Environments with Varying Complexities
}
\author{Yifei Simon Shao$^{*}$, Yuwei Wu$^{*}$, Laura Jarin-Lipschitz$^{*}$, Pratik Chaudhari, Vijay Kumar
\thanks{We gratefully acknowledge the support of The Institute for Learning-Enabled Optimization at Scale (TILOS) funded by the National Science Foundation (NSF) under NSF Grant CCR-2112665, IoT4Ag ERC funded through NSF Grant EEC-1941529, ONR grant N00014-20-1-2822, ONR grant N00014-20-S-B001, NIFA grant 2022-67021-36856.}
\thanks{$^{*}$Equal contribution. All authors are with GRASP Laboratory, University of Pennsylvania, Philadelphia, PA, 19104 USA {\tt\small\{yishao, yuweiwu, laurajar, pratikac, kumar\}@seas.upenn.edu}. }}%

\begin{document}

\maketitle

\begin{abstract}

Motion planning techniques for quadrotors have advanced significantly over the past decade. Most successful planners have two stages: a front-end that determines a path that incorporates geometric (or kinematic or input) constraints and specifies the homotopy class of the trajectory, and a back-end that optimizes this path to respect dynamics and input constraints. While there are many different choices for each stage, the eventual performance depends critically not only on these choices, but also on the environment. Given a new environment, it is difficult to decide \emph{a priori} how one should design a motion planner. In this work, we develop (i) a procedure to construct parametrized environments, (ii) metrics that characterize the difficulty of motion planning in these environments, and (iii) an open-source software stack that can be used to combine a wide variety of two-stage planners seamlessly. We perform experiments in simulations and a real platform. We find, somewhat conveniently, that geometric front-ends are sufficient for environments with varying complexities if combined with dynamics-aware backends. The metrics we designed faithfully capture the planning difficulty in a given environment. All code is available at \url{https://github.com/KumarRobotics/kr_mp_design}.

\end{abstract}
\IEEEpeerreviewmaketitle

\section{Introduction}
\begin{figure}
      \centering
      \vspace{0.1cm}
      \includegraphics[width=1\columnwidth]{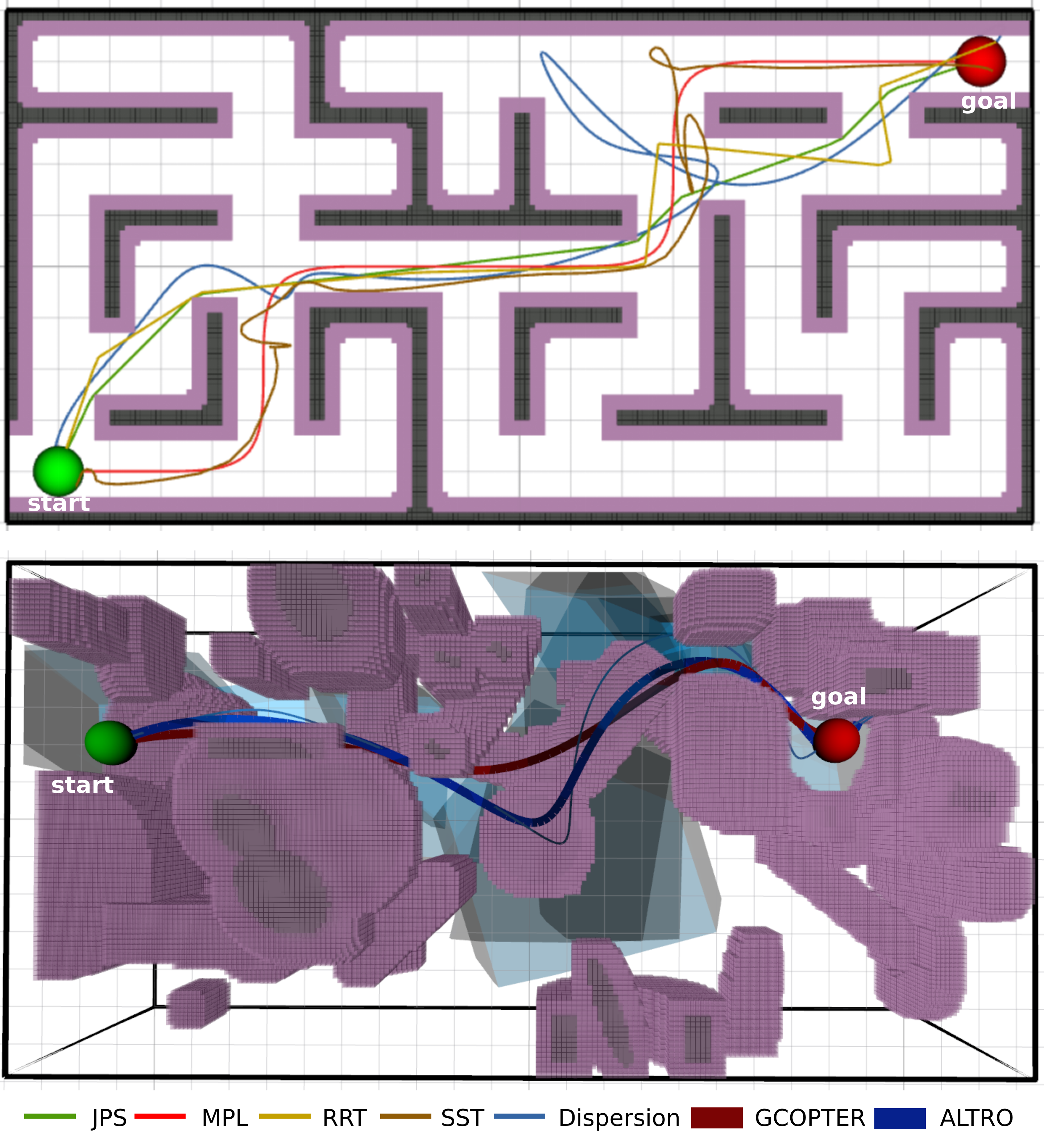}
      \caption{Overall trajectory evaluation for planning methods in different environments. The top figure demonstrates different front-end initial paths in a maze map and the bottom one shows the comparison of different back-end trajectories (thick lines) in a parameterized 3-D obstacle map. The blue polytopes are safe flight corridors.} 
      \label{fig:traj}
      \vspace{-0.4cm}
\end{figure}


Motion planning algorithms for Unmanned Aerial Vehicles (UAVs) have been extensively employed in different tasks like delivery, monitoring, and inspection in industrial and agricultural settings.
Geometric search-based and sampling-based planning methods \cite{Hart1968, Harabor2011, LaValle1998, Liu2017, li2015sparse, 9561840} can find a trajectory in a cluttered environment efficiently. 
However, finding a feasible, near-optimal, and executable trajectory for a quadrotor is nontrivial because of the complex nature of quadrotor dynamics.
Directly optimizing the feasible trajectories \cite{teng2023convex, tedrake2010lqr, marcucci2021shortest} together dramatically increases the complexity of \revise{the }problem, \revise{making it} difficult to deploy \revise{algorithms} on-board with limited computation.
Thus,  \revise{a two-stage approach \cite{2017Planning, Zhou2020e},  where a front-end algorithm provides an initial guess for the optimization and a back-end planner further refines the trajectories,  significantly improving the performance of the algorithms.}
Crucially, the performance of the back-end optimization depends on the quality and optimality of different front-end methods.

\begin{figure*}[!ht]
      \centering
       \vspace{0.1cm}
      \includegraphics[width=2.05\columnwidth]
      {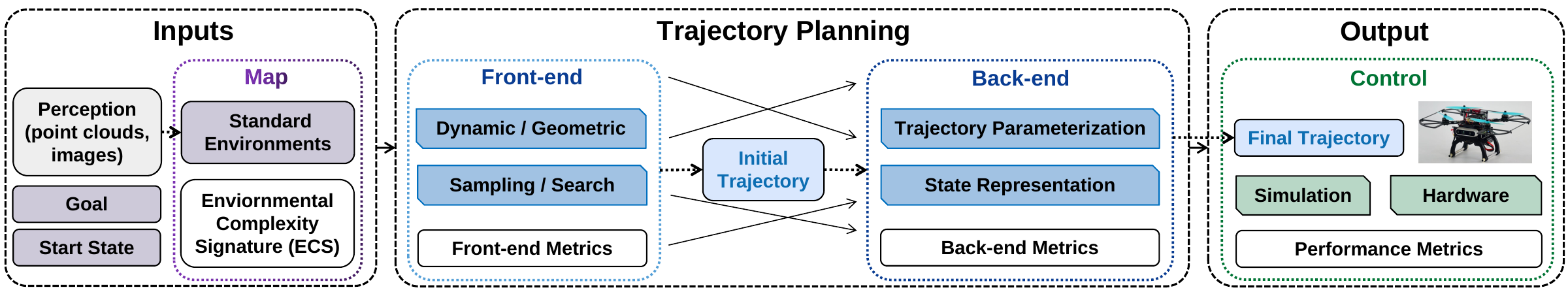}
      \caption{The architecture overview of the evaluation pipeline. The inputs for planning are the start state, goal tasks, and the standard environments from different sensors. The trajectory planning is modularized into the front-end and back-end stages, and divided in terms of its planning strategies and problem formulation. }
      \label{fig:fig1}
\end{figure*}


However, there is a lack of \revise{designing guidelines and }systemic evaluation \revise{for choosing front-end and back-end algorithms, where those guidelines and evaluation depend on the environments.}
Firstly, previous evaluations of planning algorithms are usually based on a single type of environment with a similar number of obstacles~\cite{soria2021predictive, 10160164, janson2015fast}.
More specifically, there is a lack of indoor navigation scenes, which have walls that are hard to \add{randomly generate and evaluate}, unlike the obstacles in their outdoor counterpart.
Secondly, there is a lack of modular comparison between different categories of planning methods. 
Most evaluations focus on one stage \revise{while keeping the planner of the other stage fixed, and do not provide guidelines on choosing a two-stage planner for a given environment.}
These two problems are indeed hard to solve since the same two-stage planner performs \revise{drastically differently in} different environments. 
In addition, \revise{because algorithms are often tuned for a given environment, it is hard to evaluate their performance holistically.}
Lastly, while conducting evaluations of system and hardware capabilities, the diverse sensor types and algorithms of other non-planning software introduce \revise{additional complexity making it difficult to attribute performances that are solely due to the planners.}


To \revise{address these gaps in the literature}, we propose an evaluation framework to robustly design and compare planners across a broad spectrum of environmental types.
Instead of finding the best planners, we believe we should offer consistent recommendations on choosing the most appropriate planner in different environments.
We propose \revise{the} Environment Complexity \revise{Signature (ECS)} to evaluate the environmental difficulty of different planning tasks.
To evaluate this metric and offer a parameterized map for evaluation, we open-source two new types of environments resembling indoor and outdoor settings.
We extensively conduct experiments on \revise{these} environments, along with some real-world datasets using various two-stage planners to illustrate how the environment can affect the best strategy for choosing a planner, illustrated in Fig. \ref{fig:traj}.
We developed a standardized input/output pipeline and brought in popular planners across the spectrum for evaluation.
The full pipeline is shown in Fig. \ref{fig:fig1}, where most performance metrics are obtained from a simulator. 
To validate that this metric also is grounded in reality, we conduct experiments on our hardware platform Falcon 250 v2 UAV \cite{10160295}.
The contribution of this work can be summarized as:



\begin{itemize}
 \item \revise{The first open-source modular two-stage} software stack with different planners and evaluation infrastructure.
\item \add{An} evaluation criterion \revise{called Environmental Complexity Signature (ECS)} to evaluate any given environment, \revise{which yields an estimate of} the planner performance. 
\item Two new types of parameterized point cloud generators that can generate \revise{both outdoor and indoor environments.}

\end{itemize}





\section{Related Work}\label{sec: related}
The effectiveness of planning algorithms \revise{depends on} both the environment configuration and map representations ~\cite{4650637, 10160164, 6631118}.
Environment configuration defines the characteristics of the obstacles within the planning setting. 
For instance, outdoor environments typically feature obstacles with convex geometries like cylinders and spheres, simplifying the planning process~\cite{soria2021predictive}; indoor environments often present more intricate challenges due to the non-convex obstacles~\cite{10160164, janson2015fast}. 
Previous work has analyzed planner performance in a wide array of environment configurations, ranging from forests and narrow passageways to disaster zones and urban landscapes~\cite{doi:10.1126/scirobotics.abg5810}.
Despite these efforts, there is a notable lack of a common metric to evaluate the complexities and challenges of different environmental configurations.

Not-surprisingly, the environment configuration highly influences the choice of map representation. For sparse environments with convex obstacles, geometric representations, such as ellipsoids or polytopes, are preferred due to their simplicity in both \revise{memory requirement} and constraint formulation \cite{Deits2015}. 
\revise{However}, they often assume polytopes or spherical obstacles. 
As the density of obstacles increases, geometric obstacle representation becomes unwieldy, and discrete representations, such as fields or grids, are preferred \cite{oleynikova2016signed, oleynikova2017voxblox}. 
However, in dense environments that are also structured, discrete representations may be suboptimal, since planners are prone \revise{to have solutions that are} trapped by the grids in a local minimum, particularly for reactive planners\cite{6631118}. 
The practical adoption of planning algorithms for real-world scenarios is determined by elements like the specific environment, available computational resources, and the importance of fulfilling real-time constraints.

Motion planning evaluation is frequently associated with different specific scenarios like perception tasks, autonomous navigation, and localization \cite{9645379, Montcel2019BOARRA, boroujerdian2018mavbench, yu2023avoidbench}.
\revise{For specific planning algorithms, \cite{rehberg2023comparison} evaluates the back-end minimal control trajectory optimization in scenarios of simple environments like spherical obstacles.
To systematically compare the obstacle avoidance algorithms, \cite{7759532, yu2023avoidbench} access the environments in terms of traversability and relative gap size, which are one-dimensional sampling-based metrics and cannot capture the complexity of both indoor and outdoor environments.
Indeed,} a thorough assessment of an algorithm's robustness and adaptability, should span a wide array of environmental configurations, which is not found in the literature.

\add{
Some previous work tried to compare motion planners extensively, including \cite{cohen2012generic}, focusing on mobile manipulation tasks, and \cite{moll2014extensible}, focusing on hard problems to test the limits of algorithms. However, new environments in both work need to be generated manually, making evaluation of a planner's long-term performance difficult.} 

\section{Environment Metrics} 
\label{sec: env}

\subsection{Environments}
\begin{figure}[t]
      \centering
       \vspace{0.1cm}
      \includegraphics[width=1\columnwidth]{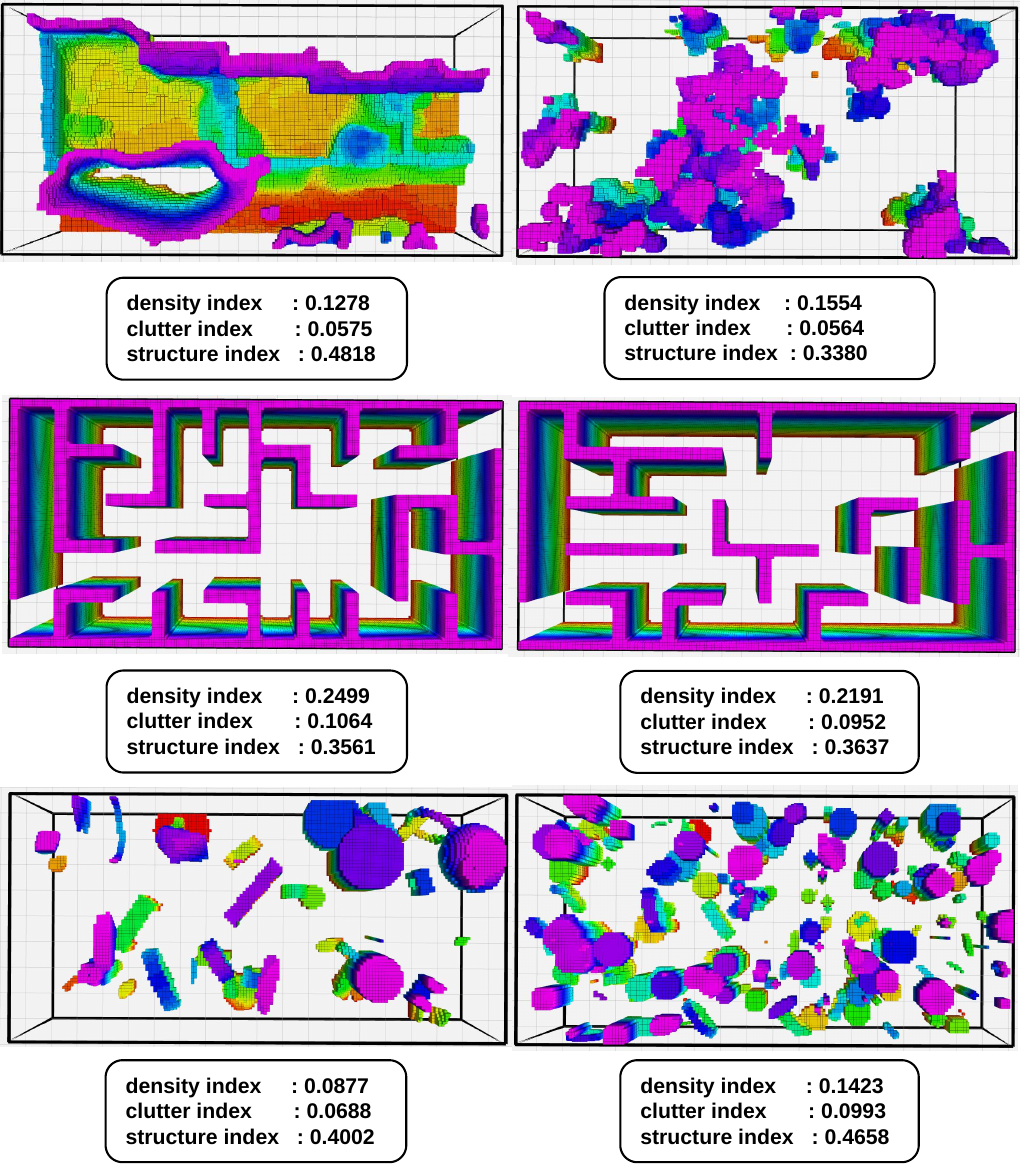}
      \caption{Examples of real maps (top), maze maps (middle), and obstacle maps (bottom) with different ECS values.}
      \label{fig:fig_env}
\end{figure}

To comprehensively validate the metrics and evaluate the environments, we provide both simulated and real-world environments, as shown in Fig. \ref{fig:fig_env}. In simulation, we provide two new types of \revise{3-D} parameterized environments:
\begin{itemize}
    \item \textit{Maze maps} are randomly generated using Kruskal's algorithm \cite{kruskal1956shortest} and parameterized by $p$, the likelihood that each wall element is deleted. We care about these maps since they resemble indoor environments, and offer different homotopy classes in 2-D for front-end planner comparisons.
    \item \textit{Obstacle maps} are generated by placing objects of different shapes into free space, which is determined by multiple geometric parameters of each type of obstacle. We select typical geometric convex objects like cylinders, ellipsoids, polytopes, and non-convex-like circles or gates to model scenarios, such as forests, warehouses, and gate racing.
\end{itemize}
\add{Besides simulated environments, we also provide} \textit{Real maps} cropped from STPLS3D \cite{Chen_2022_BMVC} and M3ED \cite{Chaney_2023_CVPR} point cloud datasets, consisting of some common environments like forests, urban cities, streets, and indoor spaces.


\subsection{Environmental Complexity Signature (ECS)}
Even though the environments we generated can be parameterized during generation, there is still a lack of a common description between all three types of environments. To bring all types of environments into a unified description, we define the Environmental Complexity \revise{Signature (ECS) with}  three metrics to distinguish the environment from sparse to dense, from dispersed to cluttered, and from unstructured to well-structured. 

\revise{Since all common sensors for perception like LiDAR or \revise{camera}-based sensors all have discrete outputs, we choose to represent an environment with 3-D points.}
For a specific quadrotor modeled as a sphere, we use its radius $r$ to scale the bounded environment $\mathcal{X} \in \mathbb{R}^3 $ with an absolute size $ s_x \times s_y \times s_z $. 
We use a grid discretization with a resolution equal to the radius of the quadrotor to approximately quantify the discretized environment  $\mathcal{X}^d $. 
The occupancy point positions we take for evaluation are the center points of grids, defined as $\mathcal{X}_{obs}^d  = \{ o_i \}_{i=1}^N \subset \mathcal{X}^d $, where \add{ $o_i\in \mathbb{R}^3$ is 3-D coordinate of $i$-th grid}, $N$ is the number of total occupied grids.

\subsubsection{Density Index} The first metric is a measure of the density of the obstacles in the environment and is given by:
\begin{equation}
    d(\mathcal{X}^d) = \frac{ r^3  \cdot  N}{ s_x \cdot   s_y  \cdot  s_z}.
\end{equation}
The index function $d(\mathcal{X}^d) \in [0, 1] $ computes the ratio of occupied grids over total grids in the bounded space. As the density index increases, the number of obstacles in the environment also increases accordingly.

\subsubsection{Clutter Index} 

Similar to the relative gap size \cite{yu2023avoidbench} to quantify the narrowest gap in the environment for the quadrotor to go through, dispersion quantifies the largest empty ball (2-norm) in the environment \cite{Lav06}, which provides the lower-bounded space to insert a quadrotor with any orientation. We use the ratio of the radius of the quadrotor scaled by dispersion to define the level of clutter in the environment,
\begin{equation}
    c(\mathcal{X}^d) = \left (\frac{1}{r} \max_{x \in \mathcal{X}_{free}^d }  \{  \min_{o_i \in \mathcal{X}^d_{obs} }    \{   \rho (x, o_i)  \}   \} \right )^{-1},
\end{equation}
where  $ \mathcal{X}_{free}^d \subset \mathcal{X} $ denotes the discretized closed free space set, 
$\rho (\cdot)$ is a metric function that the point in the free space that doesn't intersect with the occupancy points, here we use $L^2$ metric. 
The clutter index of a feasible environment should be in the range $[0, 1]$ to at least allow the quadrotor to stay in the environment, where $c = 1$ indicates that the quadrotor barely fits in the environment.

\subsubsection{Structure Index} In addition to describing the geometric complexity of the environment, we want to characterize the topological complexity which is naturally described by the number of homotopy or homology classes of trajectories between a start and a goal \cite{Munkres1974TopologyAF, bhattacharya2010search}. 
\revise{
However, the homotopy or homology classes of the whole environment are notoriously difficult to compute without traversing all the combinations of collision-free start and end goals.
Instead, we capture the topological features by using the approximated ``surface area" of obstacles. 
Obstacles with more holes (higher genus) have a larger   surface area than a solid object (zero genus).
Accordingly, we define a structure index that essentially measures qualitatively distinct paths in the environment, using}
\begin{equation}
    s(\mathcal{X}^d) = \frac{1}{N}  
  \sum_{i = 1}^N \gamma (o_i) ,
\end{equation}
where $\gamma (\cdot)$ is an indicator function that evaluates to 1 if the obstacle grid has any direct neighbor to the free space, and $s(\mathcal{X}^d) \in [0, 1] $ counts the ratio of the cell points with neighbors over all the obstacle points.

With the above metrics, we introduce a single measure of environmental complexity called the Environmental Complexity \revise{Signature (ECS)} and explore the effectiveness of different front and back-end planners.
\begin{equation}
\mathbf{ECS} = \left ( d(\mathcal{X}^d),    c(\mathcal{X}^d), s(\mathcal{X}^d) \right ).
\end{equation}
This index is independent of the quadrotor size and invariant to translations and orientations.

\section {Design and Evaluation of Motion Planners} \label{sec: planners}

\subsection{Map Representation for Planning}

The best map representation for motion planning can be different in terms of environment configurations and tasks. 
However, in order to evaluate all different planners in these environments, we employ standardized map representations for front-end planners and back-end planners.

For front-end methods, we use voxel map representation $\mathcal{M}_v(\textrm{res},\textrm{size})$ defined by its resolution and total relative ranges, which enjoys $O(1)$ collision check complexity. 
We set its resolution to half the radius of the quadrotor to ensure the efficiency and accuracy of the map. 
\revise{Given the front-end initial path, we employ general space representation as a safe flight corridor and use the method in \cite{2017Planning} to cover the path. The corridor is a series of overlapped and ordered convex polytopes $\mathcal{M}_c(\bigcup_{k = 1}^M \mathcal{P}_k)$ in the free space, and formulated as linear constraints in the back-end optimization.}

These standard map representations offer universal support for grid-based and sampling-based front-end methods and make constraint formulation easy for back-end optimization.

\subsection{Planning Algorithms}

We focus on two-staged planners for their balanced efficiency and optimality. 

The front-end planners consider different levels of fidelity and can plan in geometric space, input space, or state space, considering a quadrotor as a multi-order integrator. 
High-fidelity front-end planners can provide \revise{a more dynamically feasible initial trajectory}, which is crucial for back-end planners. The specific front-end methods arranged from low dimensional to high dimensional are
\begin{itemize}
  \item \textbf{Geometric Search}: \textit{Jump Point Search (JPS)}\cite{Harabor2011} is a improved grid search method, similar to A*, in 3-D. This method is complete and serves as the baseline for if the map is feasible.
  \item \textbf{Geometric Sampling}: \textit{RRT*}\cite{LaValle1998} in 3-D uses sampling to find a path quickly using rewiring to improve path quality with a time limit iteratively.
  \item \textbf{Input Space Search}: \textit{Motion Planning Primitive (MPL)}\cite{Liu2017} uniformly select a range of inputs in 3-D \add{(Cartesian coordinates' accelerations)} and build a graph online for planning.
  \item \textbf{Input Space Sampling}: \textit{SST (Stable Sparse RRT)}\cite{li2015sparse} randomly samples both input and input duration in 4-D to iteratively provide better solution with a time limit.
  \item \textbf{State Space Search}: \textit{Dispersion Planner}\cite{9561840} builds a graph offline in state space in 6-D \add{(2-D Cartesian coordinates positions, velocities and accelerations)}, so that a solution can be found online quickly.
\end{itemize}
The back-end planner uses first-order or second-order methods for further trajectory optimization. \revise{Hard-constraint optimization methods \cite{5980409} are firstly proposed and applied but due to the infeasibility and time allocation problems \cite{8593579}, soft-constraint optimization \cite{Zhou2020e, Wang2022} became more prevalent, and often faster.}
In these methods, \revise{there are two main categories of solutions:} those that simplify quadrotors as multi-order integrators so that differentially flatness can be exploited\cite{5980409,tordesillas2021faster}; and those that use full quadrotor dynamics for planning~\cite{sun2022comparative}. We take the \revise{state-of-art method from each category for the following comparisons, as}
\begin{itemize}
  \item \textbf{Differentially Flatness}: \textit{GCOPTER}\cite{Wang2022} uses a bilevel scheme and finds an unconstrained unique solution on the lower level and encodes constraints as a penalty term in the upper level. \add{Since time allocation between flight corridors is often considered difficult, we note that} this method also does time allocation optimization.
  \item \textbf{Full Dynamics}: \textit{ALTRO}\cite{howell2019altro} also uses a bilevel optimization that uses iLQR on the lower level for solving the problem fast, and formulate constraints as Augmented Lagrangian on the upper level. \revise{This method does not optimize the total trajectory time, and does not perform time allocation across polytopes.}
\end{itemize}

We explore whether higher fidelity front-end planning influences back-end performance. Therefore, we put the combination of these planners to test on the previously mentioned three types of environments with varying levels of difficulty.

\section{Result}
\label{sec: result}
\begin{figure*}[!h]
      \centering
            \vspace{0.1cm}
 \includegraphics[width=2.0\columnwidth]{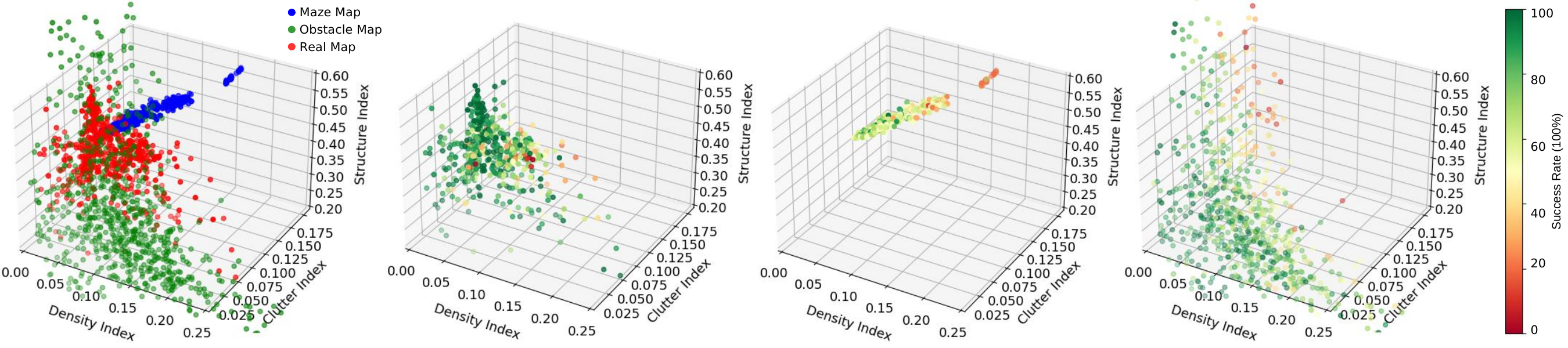}
      \caption{The visualization of the distribution of different maps. The left figure demonstrates the ECI values of maze maps, obstacle maps, and real maps. The right three figures show how many planners (out of the 10 we consider) succeeded in \revise{generating a safe trajectory} .}
      \label{fig:eci_four_scatter}
\end{figure*}

\subsection{Setups}

We generated feasible point clouds for 600 \textit{Obstacle maps}, 600 \textit{Maze maps}, and 600 \textit{Real maps}, each with varying difficulty, with the same size of $20 \ \textrm{m}  \times 10 \ \textrm{m}  \times 5 \ \textrm{m} $.
The real maps are composed of 400 cropped maps from STPLS3D \cite{Chen_2022_BMVC} with three different landscapes, and 200 cropped maps of Wharton State Forest, Pennovation outdoor and indoor scenes in M3ED \cite{Chaney_2023_CVPR}.
The quadrotor has a radius of $0.2 \ \textrm{m}$ and each environment is discretized to a voxel grid with half of the quadrotor's radius for a more accurate measurement. 
To reduce collisions caused by discretization error, we inflate the obstacles by $0.3 \ \textrm{m}$. 
we use fixed start and goal positions for each Maze map and randomize the start and goal for other maps. We filter the maps using JPS between start/goal locations, so all maps are feasible.

We conduct simulation experiments among all feasible maps for different combinations of front-end and back-end planning algorithms.
\revise{The experiments are conducted on a desktop with AMD Ryzen Threadripper 3960X, and we use ROS C++ implementations of all algorithms, RRT* and SST are from OMPL \cite{sucan2012ompl}}.
To make a fair comparison for planning algorithms, we set the same goal threshold of $1 \ m $ for all front-end planners. 
\add{Keeping real-time planning in mind, we set }the front-end planning timeout to $0.2 \ s$. We set the maximum velocity and acceleration to 3.0 $m/s$  and  2.0 $m/s^2$. The quadrotor has a mass of 1.5 kg, with a maximum thrust is 31 N. 
\revise{To ensure the accuracy of the discrete dynamics, we set ALTRO's discrete dynamics with fixed time intervals of 0.1s}.

\revise{
To evaluate the final performance of the planning stack, we evaluate \textit{success rate}, \textit{collision rate}, and \textit{computation time} of front-end planners. For back-end planners, we further evaluate \textit{average jerk} and \textit{duration of the trajectory} in addition to the front-end metrics. Lastly, to test the quality of the trajectory, we evaluate the \textit{energy cost} and \textit{tracking error} in a simulator, and verify its accuracy in the real world.}

\subsection{Comparisons in Map Metrics}

The distribution of different maps and how many planners succeeded is shown in Fig. \ref{fig:eci_four_scatter}. We consider the success percentage as the difficulty of the planning problem.
\revise{The \textit{real maps} in our datasets lie} in a smaller range of ECS than \revise{synthetically} generated \textit{obstacle maps}.
With the augmentation of \textit{obstacle maps}, we see that the clutter index is the strongest indicator of planning difficulty. 
The \textit{maze maps} with fewer walls have lower clutter indices, resembling urban \revise{\textit{real maps} in ECS and in difficulty.  }

\begin{figure}[!ht]
      \centering
      \includegraphics[width=1\columnwidth]{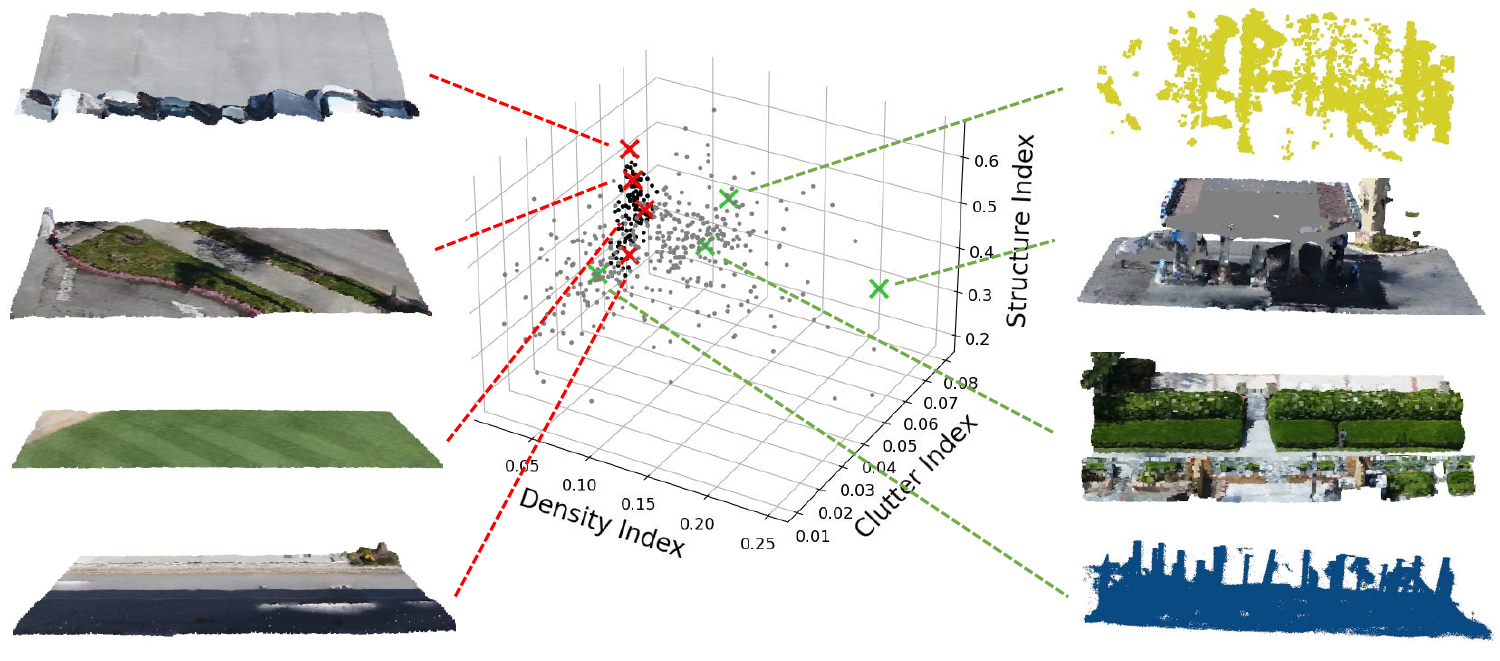}
      \caption{\revise{GMM clusters (n = 2) of \textit{real maps} from M3ED \cite{Chaney_2023_CVPR} and STPLS3D \cite{Chen_2022_BMVC} . There is a distinct cluster (black) with similar density and clutter indices values around $[0.07, 0.04]$. Random samples from each cluster are shown. } }
      \label{fig:fig_env_show}
\end{figure}

\revise{On Fig. \ref{fig:fig_env_show}, we show the scatter plot of ECS for \revise{\textit{real maps}}. There is a distinct vertical cluster of 31\% maps that has a high success rate and varies only in structure index.} From inspection, we see that they are mostly flat terrain without obstacles in the vertical direction, while the other maps are more varied. With ECS, it is clear that these maps represent a very common scenario in the dataset, and should be avoided if we want to test more challenging scenarios.

\subsection{Planning Performance}

To analyze the performance of different combinations of planners in different maps, we show the success rate result in Tab. \ref{tab: detailed_success_rate}.
The third column is the front-end success rate, and the last two columns are the total \revise{combined} success rate using different back-end planners.

\begin{table}[!htbp]
    \centering
    \renewcommand\arraystretch{1.2}
    \caption{Planning Success Rate on Different Maps. \label{tab: detailed_success_rate}}
\begin{tabular}{|l|l|l|ll|}
\hline
\multirow{2}*{Map}  &\multicolumn{2}{c|}{Frontend} & \multicolumn{2}{c|}{ Total Success} 
  \\
\cline{2-5}
 & Method & Success & GCOPTER & ALTRO  \\
\hline
\multirow{5}*{Real}  & JPS & 100.0\% &                                  80.0\%  &                                     91.6\%  \\
&RRT*   &   98.5\%       &                                   91.6\% &                                      \textbf{95.6\%} \\

&MPL    &     87.4\% &                              85.2\%  &                                      86.4\%     \\
&SST      &    73.8\%       &                                   70.7\%  &                                      61.7\%    \\
& Dispersion &  75.4\% &                                 71.3\% &                                      68.3\%             \\
\hline
\multirow{5}*{Maze}  &JPS &   100.0\% &                    87.1\%  &    75.2\%               \\
&RRT*    &  99.8\% &                   \textbf{99.1\%} &                        81.2\% \\

&MPL   &    18.7\%   &             18.7\%  &                        18.7\% \\
&SST    &   33.8\%  &    31.8\%  &                        16.9\% \\
&Dispersion  &  86.0\% &                  75.2\%  &   50.4\% \\
\hline
\multirow{5}*{Obstacle} & JPS &   100.0\% & 69.8\% &                                     89.5\% \\
& RRT*   &  99.8\% &                                    89.8\% &                                      \textbf{96.0\%} \\
& MPL   &    82.8\% &                              79.0\% &                                      74.8\%  \\
& SST    &     73.1\%       &                                                                  68.8\%  &                                      47.2\%  \\
& Dispersion&     76.5\% &                              71.5\%  &                                      55.1\% \\
\hline
\end{tabular}
\end{table}

\begin{table*}[!htbp]
    \centering
    \vspace{0.2cm}
    \renewcommand\arraystretch{1.2}
    \setlength\tabcolsep{5pt}
    \caption{ Performance of Planners for Successful Runs\label{tab: benchmark}}
\begin{tabular}{|l|c|c|c|c|c|c|c|c|c|c|c|}
\hline 
  \multicolumn{2}{|c|}{Frontend}  & \multicolumn{2}{c|}{\tabincell{c}{Total \\ Comp. Time (ms)}} &  \multicolumn{2}{c|}{\tabincell{c}{Avg. \\ Traj. Time (s)}}&   \multicolumn{2}{c|}{\tabincell{c}{Avg. \\ Traj. Jerk Sq. ($\textrm{m}^2/\textrm{s}^5$) }} & \multicolumn{2}{c|}{\tabincell{c}{Avg. \\Tracking Error (cm)}} &  \multicolumn{2}{c|}{\tabincell{c}{Avg.  \\ RPM$^3$ $\times 10^{10}$} }  \\
\cline{1-12}
 Method &  \tabincell{c}{ Comp. \\ Time (ms)}  &  GCOPTER &   ALTRO &   GCOPTER &   ALTRO &   GCOPTER &   ALTRO &  GCOPTER &   ALTRO &   GCOPTER &  ALTRO  \\
\hline 
JPS &   \textbf{77}      & 815  &  \textbf{115}   &        8.60 &        10.31  & 4.55   &  \textbf{2.99}  & 2.1 & \textbf{1.3} 
 & 7.79 & 11.13 \\
\cline{3-12}
RRT*  &  109 &479  &  135   &           8.45 &   \textbf{8.37} &    4.47 & 3.35 & 2.1 & 1.7 & \textbf{5.40} & \textbf{5.84} \\
\cline{3-12}
MPL &   184       &            \textbf{328}  &  216   &           \textbf{8.10} &  9.15 &     \textbf{4.23}  &   3.93 & 3.6 & 1.8 & 6.84 & 8.87 \\
\cline{3-12}
SST &    204 & 538 & 269  &   8.91 &    14.45 &  4.72   &  5.63 & \textbf{2.0} & 2.3 & 8.93 & 31.14  \\
\cline{3-12}
Dispersion &   185 &586  &  244 &     9.07 &  11.60  &     4.72 & 6.38 & \textbf{2.0} & 2.7 & 7.73 & 18.42 \\
\hline 
\end{tabular}
\end{table*}

For front-end planners, geometric methods like JPS and RRT* have higher success rates than dynamically feasible methods, especially in maze environments with higher clutter index.
Comparing success rate drop from frontend planner to backend planner, JPS sees more drop than RRT* because JPS plans follow the curve of obstacles, which generates a large number of corridors, making the backend problem more difficult.
Comparing back-end planners' success rates, ALTRO usually has a high success rate but underperforms GCOPTER in maze maps due to its inability to handle long sequence node points.

For successful planning iterations, qualitatively, MPL produces the best trajectories visually, while JPS and RRT* produce very similar near-optimal geometric trajectories, as shown in Fig. \ref{fig:fig1}. 
Both SST and Dispersion planners suffer from the high dimensional nature of their state space, making sub-optimal long winding trajectories. As a result, ALTRO suffers in success rate since it takes the front-end path as the reference trajectory. Contrarily, GCOPTER only utilizes the overlapped flight corridor during optimization.
A more detailed quantitative evaluation of more trajectory metrics in all maps is shown in 
Tab. \ref{tab: benchmark}. 
For large-scale planning with a longer flight corridor, GCOPTER takes a much longer computation time but outputs shorter trajectories 
than ALTRO.
As ALTRO will directly use the total trajectory time of the front-end planner, \revise{when the quality of the front-end trajectory becomes worse (SST, Dispersion), ALTRO usually finds a trajectory that is worse than GCOPTER in terms of trajectory time, jerk, and power}. 

\add{
\textbf{Hyperparameter Tuning}: For a fair comparison between methods and to achieve the best performance of all methods, we perform Bayesian hyperparameter tuning for the most sensitive parameters of each method using Optuna\cite{akiba2019optuna}. Varying the thrust-to-weight ratio from 1.5 to 2.0 also has minimal effect on the performance of the backend planners. The parameters tuned can be found on the project website.
}

With the above evaluation results, we are able to provide a practical and thoughtful guideline for choosing a planner in different environments.

\textbf{Guidelines}:  In large-scale and cluttered indoor environments (like mazes), RRT* + GCOPTER combines the best geometric initial path and a flatness-based back-end optimizer to achieve near-perfect performance. In environments with larger gaps (like urban and sparse forests), optimizing the full dynamics (ALTRO) can provide better trajectories and higher success rates, with different front-ends such as MPL to increase trajectory smoothness while minimizing computation time, and RRT*
to increase success rates and decrease trajectory time. 

\subsection{Real-world Experiments}

\begin{figure}[!ht]
      \centering
      \includegraphics[width=1\columnwidth]{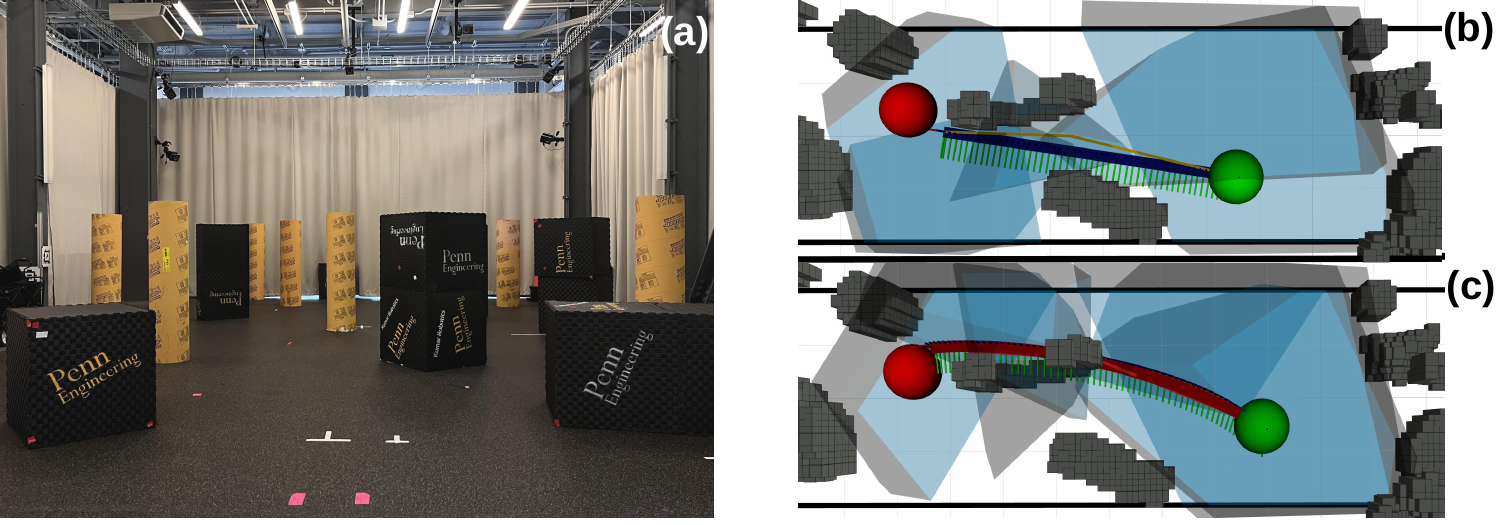}
      \caption{Experiment setting and planned trajectories. (a) The experiment environment with a motion capture system. (b) and (c) are the planned trajectories of ALTRO (dark blue) and GCOPTER (dark red) with the same front-end planner using RRT* (yellow). The green axis sequences are the executed odometry of the quadrotor. The experiment video is available at: \url{https://youtu.be/xLHHDw3IQr4}}
      \label{fig:fig_real}
\end{figure}

We use the customized hardware platform Falcon 250 v2 in \cite{10160295} to validate our evaluation pipeline in real-world scenarios. To isolate the effect of estimation and perception error, indoor experiments are conducted in a space with motion capture systems, as shown in Fig. \ref{fig:fig_real} (a). We conducted several experiments with different front-end and back-end planners and demonstrated one comparison of the final trajectories in Fig. \ref{fig:fig_real} (b, c).

\section{Conclusion} 
\revise{In this paper, we introduce a novel, modular software stack for two-stage planning algorithms. We then introduce two new types of parameterized environments along with real datasets for thorough planner evaluations. Lastly, we introduce a map evaluation criterion called ECS and make the key observations that planners should be evaluated with consideration of environmental properties. We make specific recommendations for the best two-stage planners for different environments. Interestingly, our results suggest that integrating dynamic constraints into the front-end planners may have only little to no effect over geometric front-end planners in easier environments, and has a negative effect on more difficult environments.} 

\add{For real-world navigation with an unknown environment, our approach can be extended through online ECS evaluation coupled with finite horizon planning. The best planner can be selected depending on ECS variations. We believe using the ECS is a first step for planner selection and customization for the robotics community.}

\bibliography{references}

\begin{thebibliography}{10}
\providecommand{\url}[1]{#1}
\csname url@rmstyle\endcsname
\providecommand{\newblock}{\relax}
\providecommand{\bibinfo}[2]{#2}
\providecommand\BIBentrySTDinterwordspacing{\spaceskip=0pt\relax}
\providecommand\BIBentryALTinterwordstretchfactor{4}
\providecommand\BIBentryALTinterwordspacing{\spaceskip=\fontdimen2\font plus
\BIBentryALTinterwordstretchfactor\fontdimen3\font minus \fontdimen4\font\relax}
\providecommand\BIBforeignlanguage[2]{{%
\expandafter\ifx\csname l@#1\endcsname\relax
\typeout{** WARNING: IEEEtran.bst: No hyphenation pattern has been}%
\typeout{** loaded for the language `#1'. Using the pattern for}%
\typeout{** the default language instead.}%
\else
\language=\csname l@#1\endcsname
\fi
#2}}

\bibitem{Hart1968}
P.~Hart, N.~Nilsson, and B.~Raphael, ``A formal basis for the heuristic determination of minimum cost paths,'' \emph{{IEEE} Transactions on Systems Science and Cybernetics}, vol.~4, no.~2, pp. 100--107, 1968.

\bibitem{Harabor2011}
D.~Harabor and A.~Grastien, ``Online graph pruning for pathfinding on grid maps,'' in \emph{Proceedings of the Twenty-Fifth AAAI Conference on Artificial Intelligence}, 2011, p. 1114–1119.

\bibitem{LaValle1998}
S.~M. LaValle, ``{Rapidly-Exploring Random Trees: A New Tool for Path Planning},'' Tech. Rep., 1998.

\bibitem{Liu2017}
S.~Liu, N.~Atanasov, K.~Mohta, and V.~Kumar, ``Search-based motion planning for quadrotors using linear quadratic minimum time control,'' in \emph{2017 IEEE/RSJ International Conference on Intelligent Robots and Systems (IROS)}, 2017, pp. 2872--2879.

\bibitem{li2015sparse}
Y.~Li, Z.~Littlefield, and K.~E. Bekris, ``Sparse methods for efficient asymptotically optimal kinodynamic planning,'' in \emph{Algorithmic Foundations of Robotics XI: Selected Contributions of the Eleventh International Workshop on the Algorithmic Foundations of Robotics}.\hskip 1em plus 0.5em minus 0.4em\relax Springer, 2015, pp. 263--282.

\bibitem{9561840}
L.~Jarin-Lipschitz, J.~Paulos, R.~Bjorkman, and V.~Kumar, ``Dispersion-minimizing motion primitives for search-based motion planning,'' in \emph{2021 IEEE International Conference on Robotics and Automation (ICRA)}, 2021, pp. 12\,625--12\,631.

\bibitem{teng2023convex}
S.~Teng, A.~Jasour, R.~Vasudevan, and M.~G. Jadidi, ``{Convex Geometric Motion Planning on Lie Groups via Moment Relaxation},'' in \emph{Proceedings of Robotics: Science and Systems}, Daegu, Republic of Korea, July 2023.

\bibitem{tedrake2010lqr}
R.~Tedrake, I.~R. Manchester, M.~Tobenkin, and J.~W. Roberts, ``Lqr-trees: Feedback motion planning via sums-of-squares verification,'' \emph{The International Journal of Robotics Research}, vol.~29, no.~8, pp. 1038--1052, 2010.

\bibitem{marcucci2021shortest}
T.~Marcucci, J.~Umenberger, P.~A. Parrilo, and R.~Tedrake, ``Shortest paths in graphs of convex sets,'' \emph{arXiv preprint arXiv:2101.11565}, 2021.

\bibitem{2017Planning}
S.~Liu, M.~Watterson, K.~Mohta, K.~Sun, S.~Bhattacharya, C.~J. Taylor, and V.~Kumar, ``Planning dynamically feasible trajectories for quadrotors using safe flight corridors in 3-d complex environments,'' \emph{IEEE Robotics and Automation Letters}, pp. 1--1, 2017.

\bibitem{Zhou2020e}
B.~Zhou, F.~Gao, J.~Pan, and S.~Shen, ``{Robust Real-time UAV Replanning Using Guided Gradient-based Optimization and Topological Paths},'' in \emph{Proceedings - IEEE International Conference on Robotics and Automation}, 2020, pp. 1208--1214.

\bibitem{soria2021predictive}
E.~Soria, F.~Schiano, and D.~Floreano, ``Predictive control of aerial swarms in cluttered environments,'' \emph{Nature Machine Intelligence}, vol.~3, no.~6, pp. 545--554, 2021.

\bibitem{10160164}
J.~Park, Y.~Lee, I.~Jang, and H.~J. Kim, ``Dlsc: Distributed multi-agent trajectory planning in maze-like dynamic environments using linear safe corridor,'' \emph{IEEE Transactions on Robotics}, pp. 1--20, 2023.

\bibitem{janson2015fast}
L.~Janson, E.~Schmerling, A.~Clark, and M.~Pavone, ``Fast marching tree: A fast marching sampling-based method for optimal motion planning in many dimensions,'' \emph{The International journal of robotics research}, vol.~34, no.~7, pp. 883--921, 2015.

\bibitem{10160295}
Y.~Tao, Y.~Wu, B.~Li, F.~Cladera, A.~Zhou, D.~Thakur, and V.~Kumar, ``Seer: Safe efficient exploration for aerial robots using learning to predict information gain,'' in \emph{2023 IEEE International Conference on Robotics and Automation (ICRA)}, 2023, pp. 1235--1241.

\bibitem{4650637}
K.~Yang and S.~Sukkarieh, ``3d smooth path planning for a uav in cluttered natural environments,'' in \emph{2008 IEEE/RSJ International Conference on Intelligent Robots and Systems}, 2008, pp. 794--800.

\bibitem{6631118}
D.~Bareiss and J.~van~den Berg, ``Reciprocal collision avoidance for robots with linear dynamics using lqr-obstacles,'' in \emph{2013 IEEE International Conference on Robotics and Automation}, 2013, pp. 3847--3853.

\bibitem{doi:10.1126/scirobotics.abg5810}
A.~Loquercio, E.~Kaufmann, R.~Ranftl, M.~Müller, V.~Koltun, and D.~Scaramuzza, ``Learning high-speed flight in the wild,'' \emph{Science Robotics}, vol.~6, no.~59, p. eabg5810, 2021.

\bibitem{Deits2015}
R.~Deits and R.~Tedrake, ``Efficient mixed-integer planning for uavs in cluttered environments,'' in \emph{2015 IEEE International Conference on Robotics and Automation (ICRA)}, 2015, pp. 42--49.

\bibitem{oleynikova2016signed}
H.~Oleynikova, A.~Millane, Z.~Taylor, E.~Galceran, J.~Nieto, and R.~Siegwart, ``Signed distance fields: A natural representation for both mapping and planning,'' in \emph{RSS 2016 workshop: geometry and beyond-representations, physics, and scene understanding for robotics}.\hskip 1em plus 0.5em minus 0.4em\relax University of Michigan, 2016.

\bibitem{oleynikova2017voxblox}
H.~Oleynikova, Z.~Taylor, M.~Fehr, R.~Siegwart, and J.~Nieto, ``Voxblox: Incremental 3d euclidean signed distance fields for on-board mav planning,'' in \emph{IEEE/RSJ International Conference on Intelligent Robots and Systems (IROS)}, 2017.

\bibitem{9645379}
C.~Chamzas, C.~Quintero-Peña, Z.~Kingston, A.~Orthey, D.~Rakita, M.~Gleicher, M.~Toussaint, and L.~E. Kavraki, ``Motionbenchmaker: A tool to generate and benchmark motion planning datasets,'' \emph{IEEE Robotics and Automation Letters}, vol.~7, no.~2, pp. 882--889, 2022.

\bibitem{Montcel2019BOARRA}
T.~T.~D. Montcel, A.~N{\`e}gre, J.-E. Gomez-Balderas, and N.~Marchand, ``Boarr : A benchmark for quadrotor obstacle avoidance using ros and rotors,'' \emph{ROSCon France}, 2019.

\bibitem{boroujerdian2018mavbench}
B.~Boroujerdian, H.~Genc, S.~Krishnan, W.~Cui, A.~Faust, and V.~Reddi, ``Mavbench: Micro aerial vehicle benchmarking,'' in \emph{2018 51st annual IEEE/ACM international symposium on microarchitecture (MICRO)}, 2018, pp. 894--907.

\bibitem{yu2023avoidbench}
H.~Yu, G.~C.~E. de~Croon, and C.~De~Wagter, ``Avoidbench: A high-fidelity vision-based obstacle avoidance benchmarking suite for multi-rotors,'' in \emph{2023 IEEE International Conference on Robotics and Automation (ICRA)}, 2023, pp. 9183--9189.

\bibitem{rehberg2023comparison}
W.~Rehberg, J.~Ortiz-Haro, M.~Toussaint, and W.~H{\"o}nig, ``Comparison of optimization-based methods for energy-optimal quadrotor motion planning,'' \emph{arXiv preprint arXiv:2304.14062}, 2023.

\bibitem{7759532}
C.~Nous, R.~Meertens, C.~De~Wagter, and G.~de~Croon, ``Performance evaluation in obstacle avoidance,'' in \emph{2016 IEEE/RSJ International Conference on Intelligent Robots and Systems (IROS)}, 2016, pp. 3614--3619.

\bibitem{cohen2012generic}
B.~Cohen, I.~A. {\c{S}}ucan, and S.~Chitta, ``A generic infrastructure for benchmarking motion planners,'' in \emph{2012 IEEE/RSJ International Conference on Intelligent Robots and Systems}, 2012, pp. 589--595.

\bibitem{moll2014extensible}
M.~Moll, I.~A. Sucan, and L.~E. Kavraki, ``An extensible benchmarking infrastructure for motion planning algorithms,'' \emph{arXiv preprint arXiv:1412.6673}, 2014.

\bibitem{kruskal1956shortest}
J.~B. Kruskal, ``On the shortest spanning subtree of a graph and the traveling salesman problem,'' \emph{Proceedings of the American Mathematical society}, vol.~7, no.~1, pp. 48--50, 1956.

\bibitem{Chen_2022_BMVC}
\BIBentryALTinterwordspacing
M.~Chen, Q.~Hu, Z.~Yu, H.~THOMAS, A.~Feng, Y.~Hou, K.~McCullough, F.~Ren, and L.~Soibelman, ``Stpls3d: A large-scale synthetic and real aerial photogrammetry 3d point cloud dataset,'' in \emph{33rd British Machine Vision Conference (BMVC), London, UK, November 21-24, 2022}. [Online]. Available: \url{https://bmvc2022.mpi-inf.mpg.de/0429.pdf}
\BIBentrySTDinterwordspacing

\bibitem{Chaney_2023_CVPR}
K.~Chaney, F.~Cladera, Z.~Wang, A.~Bisulco, M.~A. Hsieh, C.~Korpela, V.~Kumar, C.~J. Taylor, and K.~Daniilidis, ``M3ed: Multi-robot, multi-sensor, multi-environment event dataset,'' in \emph{Proceedings of the IEEE/CVF Conference on Computer Vision and Pattern Recognition (CVPR) Workshops}, June 2023, pp. 4015--4022.

\bibitem{Lav06}
S.~M. LaValle, \emph{Planning Algorithms}.\hskip 1em plus 0.5em minus 0.4em\relax Cambridge, U.K.: Cambridge University Press, 2006, available at http://planning.cs.uiuc.edu/.

\bibitem{Munkres1974TopologyAF}
J.~R. Munkres, ``Topology; a first course.''\hskip 1em plus 0.5em minus 0.4em\relax Prentice-Hall, 1974.

\bibitem{bhattacharya2010search}
S.~Bhattacharya, ``Search-based path planning with homotopy class constraints,'' in \emph{Proceedings of the AAAI conference on artificial intelligence}, vol.~24, no.~1, 2010, pp. 1230--1237.

\bibitem{5980409}
D.~Mellinger and V.~Kumar, ``Minimum snap trajectory generation and control for quadrotors,'' in \emph{2011 IEEE International Conference on Robotics and Automation}, 2011, pp. 2520--2525.

\bibitem{8593579}
F.~Gao, W.~Wu, J.~Pan, B.~Zhou, and S.~Shen, ``Optimal time allocation for quadrotor trajectory generation,'' in \emph{2018 IEEE/RSJ International Conference on Intelligent Robots and Systems (IROS)}, 2018, pp. 4715--4722.

\bibitem{Wang2022}
Z.~Wang, X.~Zhou, C.~Xu, and F.~Gao, ``{Geometrically Constrained Trajectory Optimization for Multicopters},'' \emph{IEEE Transactions on Robotics}, vol.~38, no.~5, pp. 3259--3278, 2022.

\bibitem{tordesillas2021faster}
J.~Tordesillas and J.~P. How, ``{FASTER}: Fast and safe trajectory planner for navigation in unknown environments,'' \emph{IEEE Transactions on Robotics}, 2021.

\bibitem{sun2022comparative}
S.~Sun, A.~Romero, P.~Foehn, E.~Kaufmann, and D.~Scaramuzza, ``A comparative study of nonlinear mpc and differential-flatness-based control for quadrotor agile flight,'' \emph{IEEE Transactions on Robotics}, vol.~38, no.~6, pp. 3357--3373, 2022.

\bibitem{howell2019altro}
T.~A. Howell, B.~E. Jackson, and Z.~Manchester, ``Altro: A fast solver for constrained trajectory optimization,'' in \emph{2019 IEEE/RSJ International Conference on Intelligent Robots and Systems (IROS)}, 2019, pp. 7674--7679.

\bibitem{sucan2012ompl}
I.~A. {\c{S}}ucan, M.~Moll, and L.~E. Kavraki, ``The {O}pen {M}otion {P}lanning {L}ibrary,'' \emph{{IEEE} Robotics \& Automation Magazine}, vol.~19, no.~4, pp. 72--82, December 2012, \url{https://ompl.kavrakilab.org}.

\bibitem{akiba2019optuna}
M.~Moll, I.~A. Sucan, and L.~E. Kavraki, ``Benchmarking motion planning algorithms: An extensible infrastructure for analysis and visualization,'' \emph{IEEE Robotics \& Automation Magazine}, vol.~22, no.~3, pp. 96--102, 2015.

\end{thebibliography}
\end{document}